\documentclass[10pt,twocolumn,letterpaper]{article}

\usepackage{iccv}
\usepackage{times}
\usepackage{epsfig}
\usepackage{graphicx}
\usepackage{amsmath}
\usepackage{amssymb}
\usepackage{multirow}
\usepackage{caption}
\usepackage{subcaption}

\usepackage{xcolor}
\newcommand\blfootnote[1]{%
  \begingroup
  \renewcommand\thefootnote{}\footnote{#1}%
  \addtocounter{footnote}{-1}%
  \endgroup
}

\usepackage[breaklinks=true,bookmarks=false]{hyperref}

\iccvfinalcopy 

\def\iccvPaperID{1784} 

\ificcvfinal\fi

\begin{document}

\title{Universal Cross-Domain Retrieval: Generalizing Across Classes and Domains}
\author{  
{Soumava Paul\textsuperscript{1}*, Titir Dutta\textsuperscript{2}*,  Soma Biswas\textsuperscript{2}}\\
{ \textsuperscript{1}Indian Institute of Technology, Kharagpur, } { \textsuperscript{2}Indian Institute of Science, Bangalore} \\
{\tt\small soumava2016@gmail.com, \{titird,somabiswas\}@iisc.ac.in}}


\maketitle
\ificcvfinal\thispagestyle{empty}\fi

\begin{abstract}
In this work, for the first time, we address the problem of universal cross-domain retrieval, where the test data can belong to classes or domains which are unseen during training.
Due to dynamically increasing number of categories and practical constraint of training on every possible domain, which requires large amounts of data, generalizing to both unseen classes and domains is important.
Towards that goal, we propose SnMpNet (Semantic Neighbourhood and Mixture Prediction Network), which incorporates two novel losses to account for the unseen classes and domains encountered during testing.
Specifically, we introduce a novel Semantic Neighborhood loss to bridge the knowledge gap between seen and unseen classes and ensure that the latent space embedding of the unseen classes is semantically meaningful with respect to its neighboring classes.
We also introduce a mix-up based supervision at image-level as well as semantic-level of the data for training with the Mixture Prediction loss, which helps in efficient retrieval when the query belongs to an unseen domain.
These losses are incorporated on the SE-ResNet50 backbone to obtain SnMpNet.
Extensive experiments on two large-scale datasets, Sketchy Extended and DomainNet, and thorough comparisons with state-of-the-art justify the effectiveness of the proposed model.

\blfootnote{* Equal contribution.}

\end{abstract}

\vspace{-4mm}

\section{Introduction}
\label{sec_introduction}
Due to the availability of large amount of data in different domains of multi-media, cross-domain retrieval has gained significant attention.
It addresses the challenging problem of retrieving relevant data from a domain~(say, image), when the query belongs to a different domain~(e.g. sketch, painting etc.).
As motivation for our work, we focus on the specific application of sketch-based image retrieval~(SBIR)~\cite{DSH}\cite{GDH}, which has wide range of applications in e-commerce, forensic data matching, etc.
Considering the dynamic real-world, where the search dataset is always being augmented with new categories of data, recently, the focus has shifted to zero-shot SBIR~(ZS-SBIR) or generalized ZS-SBIR (GZS-SBIR)~\cite{ZSIH}\cite{CVAE}\cite{Doodle}\cite{SAKE}\cite{SEM_PCYC}\cite{StyleGuide}, in which, the query and search set samples belong to classes not seen during training.

The generic architecture for ZS-SBIR (or other cross-domain retrieval applications) consists of two parallel branches, each consisting of a feature extractor and a classifier, to learn the latent-space representations of the data from individual domains (here, sketches and images). 
The domain gap in this latent space is bridged by the semantic descriptions~\cite{Word2vec}\cite{Glove} of the \emph{seen} classes. 
During testing, the query sketch and search set images are projected to this space and compared directly for retrieval.
But if the query belongs to a different domain, say painting, then this network needs to be re-trained, with painting and images as the two domains. 
This not only requires the training to be performed for every domain-pair with sufficient amount of data, but also, the query domain needs to be known a-priori.

Here, we attempt a more realistic and significantly more challenging cross-domain retrieval scenario, where the query data can belong not only to unseen classes, but also to unseen domains - which we term as universal cross-domain retrieval (UCDR).
It is a combination of two well-studied, but separate problems in literature, namely, ZS-SBIR, which accounts for test data from unseen classes and domain generalization~(DG)~\cite{eisnet}\cite{EpisodicDG}, which accounts for test data from unseen domains in the classification.
To this end, we propose \emph{SnMpNet (Semantic Neighbourhood and Mixture Prediction Network)}, which is a single-branch network consisting of a feature extractor and classifier, for learning a domain-independent embedding of input data, that also generalizes to unseen category test data. 
For generalizing to unseen classes, we propose \emph{Semantic Neighbourhood Loss}, to represent the \emph{unseen} classes in terms of their relative positions with the \emph{seen} classes.
In addition, we exploit the mix-up technique~\cite{mixup} to populate our training set with samples created through both inter-class and inter-domain mixing to prepare for unseen domains of query samples during testing.
To better generalize across domains, we propose a novel \emph{Mixture Prediction Loss}.
The contributions of this work are summarized below: \\
(1) We propose a novel framework \emph{SnMpNet}, to address the universal cross-domain retrieval scenario, where the query data may belong to \emph{seen} / \emph{unseen} classes, along with \emph{seen} / \emph{unseen} domains.
To the best of our knowledge, this is the first work in literature addressing this extremely challenging problem. \\
(2) We propose two novel losses, namely \emph{Semantic Neighbourhood} loss and \emph{Mixture Prediction} loss, to account for unseen classes and unseen domains during testing. \\
(3) Extensive experiments and analysis of the proposed framework on Sketchy-Extended~\cite{Sketchy} and DomainNet~\cite{Domainnet} datasets are reported along with other state-of-the-art approaches modified for this application. 


\section{Related Work}
\label{sec_reference}
We first describe the recent advancements in ZS-SBIR and DG, since UCDR can be considered as a combination of these.
We also discuss the recently proposed classification-protocol for samples from unseen-class and domains and explain its differences with UCDR.


\textbf{Zero-shot Sketch-based Image Retrieval~(ZS-SBIR): }
ZS-SBIR protocol was first proposed in~\cite{ZSIH}\cite{CVAE}.
Later, several algorithms~\cite{SEM_PCYC}\cite{StyleGuide}\cite{Doodle}\cite{AMDReg}\cite{progressive} have been proposed to address ZS-SBIR and its generalized version GZS-SBIR.
All these algorithms follow the standard architecture with two parallel branches.
In contrast,~\cite{SAKE} proposes a single branch of network for processing data from both domains, along with a domain-indicator to embed the domain-discriminating information.
All these algorithms use semantic-information~\cite{Word2vec}\cite{Glove} to account for the knowledge-gap, an idea which is  inspired from zero-shot learning~(ZSL)~\cite{GoodBadUgly}, which we will discuss later.
{\em UCDR generalizes the task of ZS-SBIR to additionally handle unseen domains during retrieval.}

\textbf{Domain Generalization~(DG): }
DG refers to the task of classifying data from unseen domains, when the network has been trained with data from several other domains belonging to the same classes.
This is usually addressed by learning a domain-invariant feature representation of the data, using techniques like self-supervision~\cite{self_sup_dg}, triplet loss~\cite{eisnet}, maximum mean discrepancy~(MMD)~\cite{MMD} loss, adversarial loss~\cite{adv_dg}.
Recently, meta-learning~\cite{meta_dg} and episodic training~\cite{EpisodicDG} have shown impressive performance for the DG task.
{\em UCDR generalizes DG to additionally handle unseen classes in a retrieval framework.}



\textbf{Zero-shot Domain Generalization~(ZSDG): }
Our work is also related to the well-researched Zero-shot Learning~(ZSL) task, where the goal is to classify images from unseen classes during testing. 
Several seminal works have been proposed for this problem~\cite{ESZSL}\cite{ALE}\cite{SLE}\cite{CADAVAE}\cite{RegGraZSL}\cite{GoodBadUgly}.
The knowledge gap between the seen and unseen classes is bridged using their corresponding semantic information.

Recently, few works have addressed the more realistic ZSDG task, which aims to classify unseen classes across generalized domains~\cite{Cumix}\cite{ZSDG}.
A mix-up based network is proposed in~\cite{Cumix}.
The work in~\cite{ZSDG} extends domain-generalization methods, e.g. feature-critic network~\cite{FCN}, multi-task auto-encoder~\cite{MTAE} to classify unseen-class samples by incorporating the semantic-information into their existing architecture.
Recently.~\cite{OCDVS} discusses a retrieval protocol from any source domain to any target domain, using dedicated convnets for each of the training domains.
\emph{UCDR extends the ZSDG-protocol to a retrieval framework. 
In contrast to ZSL or ZSDG protocol, no semantic information of the unseen classes are exploited in UCDR.
This makes the UCDR protocol even more realistic and challenging, since in real-world, we may not have apriori information as to which classes will be  encountered during testing.} 


\section{Problem Definition}
\label{sec_prob_statement}
First, we define the task of universal cross-domain retrieval (UCDR), and the different notations used.
We assume that labeled data from $M$-different~($M \geq 2$) domains (image, clip-art, painting, etc.) are available for training as, 
$\mathcal{D}_{train} = \underset{d \in \{1,...,D\}}{\bigcup}\{\textbf{x}_i^{c,d}, c\}_{i=1}^{N_d}$.
Here, $\textbf{x}^{c,d}_i$ is the $i^{th}$ sample from $d^{th}$-domain, which belongs to $c^{th}$ class.
$N_d$ is the number of examples in the $d^{th}$ domain.
Clearly, $M=2$ represents the training set for standard cross-domain retrieval.
The class labels $c$ for all the domains belong to \emph{seen} class set $\mathcal{C}_{train}$. 
The goal is to find a latent domain-independent subspace, $\Phi \subset \mathbb{R}^m$, such that, samples from the same class across all domains come closer and samples from different classes are pushed away in this space. 
Thus, for a query $\mathbf{x}_{q}$ and a search set $\mathcal{D}_{s} = \{\mathbf{x}_s\}_{s=1}^{N_s}$, we can retrieve the relevant search-set data using nearest neighbour of the query sample, projected in this learned $\Phi$-space.

The proposed UCDR protocol is a combination of two separate experimental frameworks, namely: 
1) U$^{c}$CDR - where the query $\mathbf{x}_q$ belongs to an \emph{unseen} class, but seen domain $d \in \{1,...,D\}$.
This implies that $\mathcal{C}_{train} \cap \mathcal{C}_{test} = \phi$, where $\mathcal{C}_{test}$ is the set of possible classes of $\mathbf{x}_q$;
2) U$^{d}$CDR - where the domain of $\mathbf{x}_q$ is \emph{unseen}, but the class is \emph{seen}, i.e., $d \notin \{1,...,D\}$, but $\mathcal{C}_{test} = \mathcal{C}_{train}$. 
The proposed combined  protocol, where both the classes and domains of $\mathbf{x}_q$ can be \emph{unseen} is denoted as U$^{c,d}$CDR, or simply UCDR~(to avoid notational clutter).
ZS-SBIR is a special case of U$^{c}$CDR, where sketch and real-images are the two domains.
Also, U$^{d}$CDR is extension of DG protocol towards retrieval.




\section{Proposed Approach}
\label{sec_proposed}
\begin{figure*}[ht!]
\centering
\includegraphics[width=\textwidth]{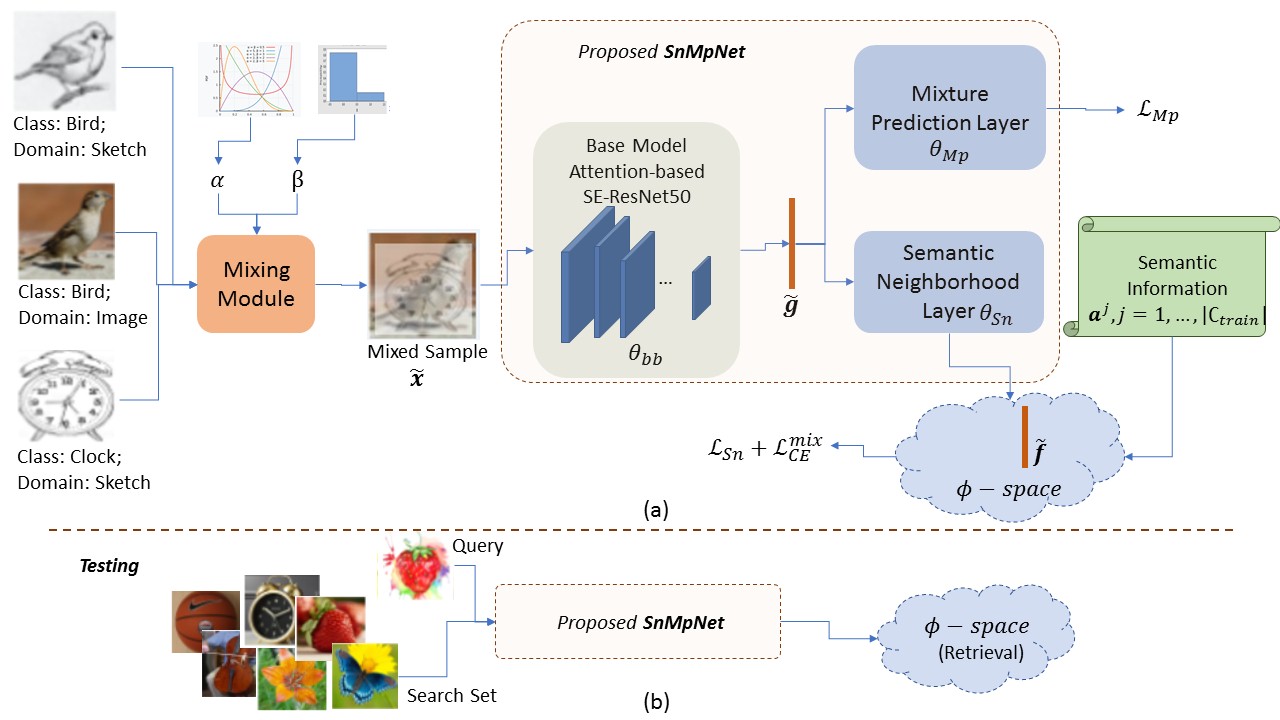}
\caption{Depiction of the proposed \emph{SnMpNet}: (a) illustrates the training methodology using mix-up and customized \emph{Mixture Prediction layer} and \emph{Semantic Neighbourhood layer} on top of the \emph{Base Model}; (b) illustrates the testing under proposed UCDR protocol, where query samples during retrieval can come from \emph{unseen domain} and \emph{unseen category}.}
\label{fig_network}
\end{figure*}

Here, we describe the proposed framework {\bf SnMpNet} in details for addressing the UCDR task.
SnMpNet is a single branch network consisting of a feature-extractor and a classifier.
Our main contributions are the \emph{semantic neighbourhood loss} to account for unseen classes, and \emph{mixture prediction loss} to account for unseen domains, integrated with a base network. \\ \\
{\bf Proposed SnMpNet Framework - Overview:} The proposed architecture for SnMpNet is illustrated in Figure~\ref{fig_network}.
For this work, we choose SE-ResNet50~\cite{SE_Resnet} as the backbone module for SnMpNet, motivated by its state-of-the-art performance for ZS-SBIR task~\cite{SAKE}.
Additionally, we incorporate an attention mechanism on top of this backbone, as in~\cite{Doodle}.
The embedding obtained for the input sample, $\mathbf{x}_{i}^{c,d}$ from the base network is denoted as $\mathbf{g}^{c,d}_i = \theta_{bb}(\mathbf{x}_i^{c,d})$.
This embedding is passed through the linear \emph{Mixture Prediction} layer $\theta_{Mp}$, which ensures that $\mathbf{g}_{i}^{c,d}$ is domain-invariant.
Next, this domain-invariant feature is passed through the linear \emph{Semantic Neighbourhood} layer, $\theta_{Sn}$ to obtain the $m$-dimensional latent space representation $\mathbf{f}_i^{c,d} = \theta_{Sn}(\mathbf{g}_i^{c,d}) \in \mathbb{R}^m$.
This $m$-dimensional space is the latent space, $\Phi$, where we obtain semantically meaningful domain-independent representations of the data and effectively perform retrieval during testing.

The learning of this $\Phi$-space is driven by two objectives: 
1) Unseen-class representation: We want to represent data from unseen classes (during testing) in a semantically meaningful manner in this space, taking into account the neighbourhood information. This is handled by the \emph{Semantic Neighborhood loss}~($\mathcal{L}_{Sn}$).
2) Domain-independent representation: We want the $\Phi$-space representation to be independent of the domain of the input data, so that SnMpNet can accommodate data from unseen domains. This is addressed by the \emph{Mixture Prediction loss}~($\mathcal{L}_{Mp}$). 
We further incorporate an inter-class and inter-domain mix-up, to generate mixed samples $\tilde{\mathbf{x}}$ and maintain only the categorical discrimination in $\Phi$-space by minimizing the \emph{Mixup Classification} loss~($\mathcal{L}_{CE}^{mix}$).
Next, we describe the individual loss components to address the above objectives.
\subsection{Unseen-class representation}
\label{subsec_mse_loss_description}
The main challenge in handling unseen classes is to effectively and meaningfully embed them in the latent feature space $\Phi$, without any prior knowledge of those classes.
Here, we propose to learn the $\Phi$-space embeddings of the training samples, so that they are semantically meaningful with respect to the other \emph{seen}-classes, especially its neighbour classes.
Thus, during testing, the model learns to embed the unseen class query samples according to their semantic relevance into the $\Phi$-space.
This is in contrast to the cross-entropy loss used for classification, or the standard metric learning losses, like triplet loss used for retrieval, where the goal is to bring data from the same class closer and move those from other classes far apart.
Previous attempts to include the neighbourhood information in the embedding space has been reported in~\emph{Stochastic Neighbourhood Embedding}~\cite{tsne}, \emph{Memory-based Neighbourhood Embedding}~\cite{MNE} etc.
Here, we propose a novel \emph{Semantic Neighbourhood} loss for this task as described next.

We learn the feature $\mathbf{f}_i^{c,d}$, such that its distance with respect to the seen classes is same as the distance between its class-semantics and the semantics of the other seen classes.
Additionally, we introduce a strict-to-relaxed penalty term for enforcing this constraint, which depends on the semantic distance of class-$c$ to the other seen classes.
Formally, the \emph{Semantic Neighbourhood} loss is given by
\begin{align}
\label{loss_Sn}
    \mathcal{L}_{Sn} &= \sum_{\mathbf{x_i}^{c,d} \in \mathcal{D}_{train}}{\mathbf{w}(c) \odot ||\mathbb{D}(\mathbf{f}_i^{c,d}) - \mathbb{D}_{gt}(\mathbf{f}_i^{c,d})||^2}
\end{align}
where $\mathbb{D}(\mathbf{f}_i^{c,d}) \in \mathbb{R}^{|\mathcal{C}_{train}|}$, such that, its $j^{th}$ element contains the Euclidean distance between $\mathbf{f}_i^{c,d}$ and the semantic-information of the $j^{th}$ class, denoted by $\mathbf{a}^j$.
Similarly, the $j^{th}$ element of the corresponding ground-truth distance-vector ( $\mathbb{D}_{gt}(\mathbf{f}_i^{c,d})$) is the Euclidean distance between $\mathbf{a}^c$ and $\mathbf{a}^j$.
$\odot$ represents element-wise multiplication.
The $j^{th}$ entry of the weight vector $\mathbf{w}(c) \in \mathbb{R}^{|\mathcal{C}_{train}|}$ is
\begin{align}
\label{eq_wtvec_defn}
    \mathbf{w}(c)_j &= exp^{-\kappa D_{n} (\mathbf{a}^c, \mathbf{a}^j)}
\end{align}
where, $D_n(\mathbf{a}^c, \mathbf{a}^j) = \frac{D(\mathbf{a^c},\mathbf{a}^j)}{\underset{k, k \in \mathcal{C}_{train}}{max} D(\mathbf{a}^c, \mathbf{a}^k)}$; $D(.,.)$ represents the Euclidean distance between two vectors.
$\kappa$ is an experimental hyper-parameter, set using validation set accuracy.
For a smaller $D_n(\mathbf{a}^c, \mathbf{a}^j)$~(semantically similar classes), $\mathbf{w}(c)_j$ is higher, which enforces greater emphasis to preserve the relative distance for the similar classes as desired. 
For distant classes, this constraint is less strictly enforced.

\subsection{Domain-independent representation}
\label{subsec_domind_rep}
Here, we propose to learn the data-representation by incorporating mix-up based supervision, such that SnMpNet only learns the class-information of the mixed sample and obtains a domain-invariant embedding. \\
\textbf{Mixing up input data:}
Inspired by~\cite{Cumix}, we mix samples from multiple classes, as well as from multiple domains to form the set, $\mathcal{D}_{mixup} = \{ \tilde{\mathbf{x}} \}$, such that
\begin{align}
\label{eq_mixing}
    \tilde{\mathbf{x}} = \alpha \mathbf{x}_i^{c,d} + (1-\alpha)[\beta \mathbf{x}_j^{p,d} + (1-\beta)\mathbf{x}_k^{r,n}]
\end{align}
where $\alpha \sim Beta(\lambda, \lambda)$ and $\beta \sim Bernoulli(\gamma, \gamma)$,
$\lambda$ and $\gamma$ being hyper-parameters.
Clearly, $\beta=1$ results in intra-domain and $\beta=0$ results in cross-domain mix-up.
We use the samples in $\mathcal{D}_{mixup}$ for training. \\ \\
\textbf{Mixture Prediction Loss:}
We aim to remove any domain-related information from $\tilde{\mathbf{g}} = \theta_{bb}(\tilde{\mathbf{x}})$ by means of such cross-domain mixture samples.
For this, we propose a novel \emph{Mixture Prediction loss}, where the network is trained to predict the exact proportion of the component categories in a sample $\tilde{\mathbf{x}}$, and forgets about its component domains.
We incorporate this constraint before learning the $\Phi$-space representation, to ensure that the \emph{Semantic Neighbourhood loss} does not get hindered by any domain knowledge.

For this, we compute the logit-vectors of $\tilde{\mathbf{g}}$, 
by passing it through the mixture prediction layer.
The softmax activation on the $j^{th}$ element of $\theta_{Mp}(\tilde{\mathbf{g}})$ can be interpreted as the probability that sample $\tilde{\mathbf{x}}$ belongs to class-$j$:
\begin{align}
\label{eq_logit_prob}
    Prob(\tilde{\mathbf{x}} \in Class-j) &= \frac{exp(\theta_{Mp}(\tilde{\mathbf{g}})_j)}{\sum_{t \in \mathcal{C}_{train}}{exp(\theta_{Mp}(\tilde{\mathbf{g}})_t)}}
\end{align}
where $\theta_{Mp}(\tilde{\mathbf{g}})_j$ is the logit-score obtained at $j^{th}$ index.
However, $\tilde{\mathbf{g}}$ contains characteristics from its component classes, according to the mixing coefficients.
We propose to predict the mixing coefficients of those component classes through a soft cross-entropy loss designed as:
\begin{align}
    \mathcal{L}_{Mp} &= \sum_{\tilde{\mathbf{x}} \in \mathcal{D}_{mixup}}{\sum_{t=1}^{|\mathcal{C}_{train}|}{-\tilde{\mathbf{l}}_{t} \log Prob(\tilde{x} \in Class-t)}}
\end{align}
where, $\tilde{\mathbf{l}}_t$ is $t^{th}$-element of a $|\mathcal{C}_{train}|$-dimensional vector $\tilde{\mathbf{l}}$, with the mixing coefficients at their corresponding class-indices and zero at other places, e.g, for $\tilde{\mathbf{x}}$ \color{black} generated through $\beta=1$ in~\eqref{eq_mixing}:
\[
    \tilde{\mathbf{l}}_t= 
\begin{cases}
    \alpha,& \text{if } t=c\\
    (1-\alpha), & \text{if }t=p\\
    0, & \text{otherwise}
\end{cases}
\]
\color{black}
Thus, the network only remembers the cross-category mixing proportions and is independent of the input domains. \\ \\
\textbf{Mix-up Classification loss:}
Finally, we impose the standard cross-entropy loss to ensure that the discriminability of feature-space is preserved.
For this, we classify the latent $\Phi$-space representation ($\tilde{\mathbf{f}} \in \mathbb{R}^d$) of the mixed-up sample $\tilde{\mathbf{x}}$ into its component classes as in~\cite{mixup}\cite{Cumix}.
Additionally, we also want to maintain a meaningful semantic structure of the latent space, to make provision for unseen classes.
Here, we utilize the semantic information of the seen classes to address both the stated requirements.
Towards this goal, we compute the logit-score for feature $\tilde{\mathbf{f}}$ as $\mathbf{s}(\tilde{\mathbf{f}}) \in \mathbb{R}^{|\mathcal{C}_{train}|}$, so that its $j^{th}$ element can be expressed as
\begin{align}
    \mathbf{s}(\tilde{\mathbf{f}})_j &= \frac{exp(\text{cosine-similarity}(\tilde{\mathbf{f}}, \mathbf{a}^j))}{\sum_{t \in \mathcal{C}_{train}}{exp(\text{cosine-similarity}(\tilde{\mathbf{f}}, \mathbf{a}^t))}} 
\end{align}
Similar to~\eqref{eq_logit_prob}, $\mathbf{s}(\tilde{\mathbf{f}})_j$ also represents the probability that  $\tilde{\mathbf{f}}$ belongs to $j^{th}$-class.
Now, if $\tilde{\mathbf{x}}$ came from a particular class, classification can be done by minimizing the following
\begin{align}
\label{eq_standard_CE}
    \mathcal{L}_{CE}(\mathbf{y}(\tilde{\mathbf{x}}),\mathbf{s}(\tilde{\mathbf{f}})) & = \sum_{\tilde{\mathbf{x}} \in \mathcal{D}_{mixup}}{ \sum_{t=1}^{|\mathcal{C}_{train}|}{- \mathbf{y}(\tilde{\mathbf{x}})_t \log \mathbf{s}(\tilde{\mathbf{f}})_t}}
\end{align}
where, $\mathbf{y}(\tilde{\mathbf{x}})_t$ is the $t^{th}$ element of the one-hot representation of $\tilde{\mathbf{x}}$'s ground-truth class.
Since the input $\tilde{\mathbf{x}}$ does not belong to a single class, we cannot directly use such computation.
\color{black} Instead, we extend equation~\eqref{eq_standard_CE} to accommodate all the component classes of $\tilde{\mathbf{x}}$ as follows,
\begin{align}
\label{eq_mix_CE}
\mathcal{L}_{CE}^{mix} =& \alpha \mathcal{L}_{CE}(\mathbf{y}(\mathbf{x}_i^{c,d}),\mathbf{s}(\tilde{\mathbf{f}})) \nonumber \\
& + (1-\alpha)\mathcal{L}_{CE}([\beta \space \mathbf{y}(\mathbf{x}_j^{p,d}) + (1-\beta) \mathbf{y}(\mathbf{x}_k^{r,n})], \mathbf{s}(\tilde{\mathbf{f}})) 
\end{align}
\color{black}
\subsection{Combined Loss Function}
Finally, to account for both unseen classes and unseen domains during retrieval, we combine the advantages of both the above representations seamlessly in the proposed framework.
To incorporate the effect of inter-class and inter-domain mix-up in the \emph{Semantic Neighbourhood loss}, we compute $\mathbb{D}(\tilde{\mathbf{f}})$ and $\mathbb{D}_{gt}(\tilde{\mathbf{f}})$ instead of  $\mathbb{D}(\mathbf{f}_i^{c,d})$ and $\mathbb{D}_{gt}(\mathbf{f}_i^{c,d})$ in~\eqref{loss_Sn}.
\color{black}
In addition, we evaluate the mixed-up semantic information of $\tilde{\mathbf{x}}$ as the combination of its component classes in appropriate ratio as,
\begin{align}
\tilde{\mathbf{a}} &= \alpha \mathbf{a}^c + (1-\alpha)[\beta \mathbf{a}^p+(1-\beta) \mathbf{a}^r]
\end{align}
\color{black}
This modification reflects in evaluation of $\mathbb{D}_{gt}(\tilde{\mathbf{f}})$ as its $j^{th}$ component becomes the Euclidean distance between $\tilde{\mathbf{a}}$ and $\mathbf{a}^j$.
With this modification, the mix-up based supervision is introduced not only at the image-level, but also in the semantic information level.
Combining all the loss components, the final loss to train the model is
\begin{align}
\mathcal{L} &= \mathcal{L}_{CE}^{mix} + \gamma_1\mathcal{L}_{Mp} + \gamma_2\mathcal{L}_{Sn}
\end{align}
where $\gamma_1$ and $\gamma_2$ are experimental hyper-parameters to balance the contribution of the different loss components.
\begin{table*}[ht!]
\footnotesize
\begin{center}
\begin{tabular}{|c|c|c|c|c|c|}
\hline
\multicolumn{2}{|c|}{Method} & Backbone network & output dim. & mAP@200 & Prec@200 \\
\hline
\multirow{4}{*}{Existing SOTA} & CVAE~\cite{CVAE}~(ECCV, 2018) & VGG-16 & 1024 & 0.225 & 0.333 \\
& Doodle-to-Search~\cite{Doodle}~(CVPR, 2019) & VGG-16 & 300 & 0.4606 & 0.3704 \\
& SAKE-512~\cite{SAKE}~(ICCV, 2019) & SE-ResNet50 & 512 & 0.497 & 0.598 \\
& SAKE-512 (our evaluation) & SE-ResNet50 & 512 & 0.6246 & 0.5518 \\
\hline
\multirow{4}{8em}{Doodle-to-search variants} & Doodle-SingleNet & VGG-16 & 300 & 0.3743 & 0.3308 \\
& Doodle-SingleNet-w/o Label* & VGG-16 & 300 & 0.3726 & 0.3233 \\
& Doodle-SE-SingleNet & SE-ResNet50 & 300 & 0.4022 & 0.3595 \\
& Doodle-SE-SingleNet-w/o Label* & SE-ResNet50 & 300 & 0.3980 & 0.3508\\
\hline
\multirow{2}{*}{SAKE-variants} & SAKE-512-w/o Label* & SE-ResNet50 & 512 & 0.5484 &  0.4880 \\
& SAKE-300-w/o Label* & SE-ResNet50 & 300 & 0.5192 &  0.4605 \\
\hline
\multicolumn{2}{|c|}{\textbf{\emph{SnMpNet}}} & SE-ResNet50 & 300 & \textbf{0.5781} & \textbf{0.5155} \\
\hline
\end{tabular}
\end{center}
\vspace{-4mm}
\caption{Comparison for ZS-SBIR on Sketchy extended~\cite{CVAE}. Methods marked with '*' can potentially be used for UCDR. }
\label{tab_sketchy}
\end{table*}
\subsection{Retrieval}
\label{sec_retrieval}
During retrieval, for any query data $\mathbf{x}_q$, we extract the latent space representation $\mathbf{f}_q \in \mathbb{R}^m$ using the trained model.
Similarly, we also extract the latent representations of the search set samples $\mathbf{x}_s \in \mathcal{D}_{s}$ as $\mathbf{f}_s$.
We use Euclidean distance between $\mathbf{f}_q$ and $\mathbf{f}_s, s= 1,...,|\mathcal{D}_{s}|$ to rank the search set images in the final retrieval list. \\ \\
{\bf Implementation Details:}
\label{sec_implementation}
We use PyTorch 1.1.0 and a single GeForce RTX 2080 Ti GPU for implementation.
Models are trained for a maximum of $100$ epochs with early stopping of $15$ epochs based on the validation set performance.
We use SGD with nesterov momentum of $0.9$, and a batch size of $60$ to solve the optimization problem, with an initial learning rate of 1e-3, decayed exponentially to 1e-6 in $20$ epochs.
$300$-d GloVe~\cite{Glove}-embeddings and
L2-normalized \emph{word2vec} embeddings~($300$-d)~\cite{Cumix} are used as the semantic information for Sketchy and DomainNet respectively.
The key hyperparameters of \emph{SnMpNet} are $\kappa$, $\gamma_1$ and $\gamma_2$, which are set as
$\kappa\in \{1, 2\}$, $\gamma_1\in \{0.5, 1\}$, and $\gamma_2 = 1$ for both the datasets.

\section{Experiments}
\label{sec_experiment}
We now present the experimental evaluation for the proposed SnMpNet.
To the best of our knowledge, this is the first work addressing UCDR, thus there are no established baselines for direct comparison.
First, we analyze SnMpNet for U$^c$CDR protocol, where only the classes are unseen during retrieval.
We specifically consider the application of ZS-SBIR, which is well-explored in literature, to directly compare SnMpNet with current SOTA in ZS-SBIR.
Then, we extend our evaluation to the completely general UCDR setting.
We start with a brief introduction to the datasets. \\ \\
{\bf Datasets Used:} We use two datasets for the experiments. 
\textbf{Sketchy extended~\cite{Sketchy}} contains $75,471$ sketches and $73,002$ images from $125$ categories and is used for U$^c$CDR. 
To obtain completely \emph{unseen} test classes~(if pre-trained backbones are used), we follow the split in~\cite{CVAE} and  consider $21$-classes~(not part of ImageNet-1K) to be \emph{unseen}.
Among rest $104$ \emph{seen} classes, following~\cite{Doodle}, $93$ and $11$-classes are used for training and validation respectively.
\textbf{DomainNet~\cite{Domainnet}} has approximately $6,00,000$ samples from $345$ categories, collected in six domains, namely, \emph{Clip-art}, \emph{Sketch}, \emph{Real}, \emph{Quickdraw}, \emph{Infograph}, and \emph{Painting} and is used for U$^d$CDR and UCDR experiments. 
Following~\cite{Cumix}, the test set is formed with $45$ \emph{unseen} classes.
Rest $245$ and $55$ classes are used for training and validation~\cite{Cumix}.
In addition, we leave one domain~(randomly selected) out while training, to create \emph{unseen}-domain query.
The search-set is constructed with \emph{Real} images from \emph{seen} and/or \emph{unseen} classes.

\begin{table*}[ht!]
\footnotesize
\begin{center}
\begin{tabular}{|c|c|c|c|c|c|c|}
\hline
Training & Query & \multirow{2}{*}{Method} & \multicolumn{2}{c|}{\emph{Unseen}-class Search Set} & \multicolumn{2}{c|}{\emph{Seen}+\emph{Unseen}-class Search Set} \\
\cline{4-7}
Domains & Domain &  & mAP@200 & Prec@200 & mAP@200 & Prec@200\\
\hline
\emph{Real}, \emph{Quickdraw} & \multirow{3}{*}{\emph{Sketch}} & EISNet-retrieval & 0.2611 & 0.2061 & 0.2286 & 0.1805 \\
\emph{Infograph}, \emph{Painting} & & CuMix-retrieval & 0.2736 & 0.2168  & 0.2428 & 0.1935 \\
\emph{Clip-art} & & \textbf{\emph{SnMpNet}} & \textbf{0.3007}  & \textbf{0.2432} & \textbf{0.2624} & \textbf{0.2134}\\
\hline 
\emph{Real}, \emph{Sketch} & \multirow{3}{*}{\emph{Quickdraw}} & EISNet-retrieval & 0.1273 & 0.1016 & 0.1101 & 0.0870\\
\emph{Infograph}, \emph{Painting} &  & CuMix-retrieval & 0.1304 & 0.1006 & 0.1118 & 0.0852\\
\emph{Clip-art} &  & \textbf{\emph{SnMpNet}} & \textbf{0.1736} & \textbf{0.1284} & \textbf{0.1512} & \textbf{0.1111}\\
\hline
\emph{Real}, \emph{Sketch} & \multirow{3}{*}{\emph{Painting}} & EISNet-retrieval & 0.3599 & 0.2913 & 0.3280 & 0.2653\\
\emph{Infograph}, \emph{Quickdraw} &  & CuMix-retrieval & 0.3710 & 0.3001 & 0.3400 & 0.2751\\
\emph{Clip-art} &  & \textbf{\emph{SnMpNet}} & \textbf{0.4031} & \textbf{0.3332} & \textbf{0.3635} & \textbf{0.3019}\\
\hline
\emph{Real}, \emph{Sketch} & \multirow{3}{*}{\emph{Infograph}} & EISNet-retrieval & 0.1878 & 0.1512 & 0.1658 & 0.1323\\
\emph{Painting}, \emph{Quickdraw} &  & CuMix-retrieval & 0.1931 & 0.1543 & 0.1711 & 0.1361\\
\emph{Clip-art} &  & \textbf{\emph{SnMpNet}} & \textbf{0.2079} & \textbf{0.1717} & \textbf{0.1800} & \textbf{0.1496}\\
\hline
\emph{Real}, \emph{Sketch} & \multirow{3}{*}{\emph{Clip-art}} & EISNet-retrieval & 0.3585 & 0.2792 & 0.3251 & 0.2496\\
\emph{Painting}, \emph{Quickdraw} &  & CuMix-retrieval & 0.3764 & 0.2911 & 0.3428 & 0.2627\\
\emph{Infograph} &  & \textbf{\emph{SnMpNet}} & \textbf{0.4198} & \textbf{0.3323} & \textbf{0.3765} & \textbf{0.2959}\\
\hline
\multicolumn{2}{|c|}{\multirow{3}{*}{\textbf{\emph{Average}}}} & EISNet-retrieval & 0.2589 & 0.2059 & 0.2315 & 0.1829\\
\multicolumn{2}{|c|}{} & CuMix-retrieval & 0.2689 & 0.2126  & 0.2417 & 0.1905\\
\multicolumn{2}{|c|}{} & \textbf{\emph{SnMpNet}} & \textbf{0.3010}  & \textbf{0.2418} & \textbf{0.2667} & \textbf{0.2144}\\
\hline
\end{tabular}
\end{center}
\vspace{-4mm}
\caption{UCDR evaluation results on DomainNet for two different scenarios, when the search set contains (1) only \emph{unseen}-class image samples, and (2) both \emph{seen} and \emph{unseen} class samples.}
\label{tab_UCDR}
\end{table*}

\begin{table*}[ht!]
\footnotesize
\begin{center}
\begin{tabular}{|c|c|c|c|c|c|c|}
\hline
Training & Query & \multirow{2}{*}{Method} & \multicolumn{2}{c|}{\emph{Unseen}-class Search Set} & \multicolumn{2}{c|}{\emph{Seen}+\emph{Unseen}-class Search Set} \\
\cline{4-7}
Domains & Domain &  & mAP@200 & Prec@200 & mAP@200 & Prec@200\\
\hline
\emph{Real}, \emph{Quickdraw} & \multirow{3}{*}{\emph{QuickDraw}} & EISNet-retrieval & 0.2475 & 0.1906 & 0.2118 & 0.1627 \\
\emph{Infograph}, \emph{Painting} & & CuMix-retrieval & 0.2546 & 0.1967  & 0.2177 & 0.1699 \\
\emph{Clip-art} & & \textbf{\emph{SnMpNet}} & \textbf{0.2888}  & \textbf{0.2314} & \textbf{0.2366} & \textbf{0.1918}\\
\hline 
\emph{Real}, \emph{Sketch} & \multirow{3}{*}{\emph{Sketch}}  & EISNet-retrieval & 0.3719 & 0.3136 & 0.3355 & 0.2822\\
\emph{Infograph}, \emph{Painting} &  & CuMix-retrieval & 0.3689 & 0.3069 & 0.3300 & 0.2714\\
\emph{Clip-art} &  & \textbf{\emph{SnMpNet}} & \textbf{0.4221} & \textbf{0.3496} & \textbf{0.3767} & \textbf{0.3109}\\
\hline 
\end{tabular}
\end{center}
\vspace{-4mm}
\caption{U$^c$CDR-evaluation results on DomainNet for two different scenarios, when the search set contains (1) only \emph{unseen}-class image samples, and (2) both \emph{seen} and \emph{unseen} class samples.}
\label{tab_UcCDR}
\vspace{-5mm}
\end{table*}
\subsection{U$^c$CDR Evaluation}
We first analyze SnMpNet for U$^c$CDR, specifically, ZS-SBIR, where the query domain is sketch and the search set contains images, both from a set of classes unseen to the model.
Here, we train SnMpNet with sketch and image data, following the same training protocol as in~\cite{Doodle}\cite{CVAE}.\\ 
\textbf{Baseline Methods:}
First, we discuss the baseline methods and their modifications used for fair comparison.
Specifically, we develop variants with no access to domain-label of data, so that they can handle unseen domain query data.
\begin{enumerate}
    \item \textbf{Doodle-to-Search~\cite{Doodle}} trains two parallel VGG-16 networks with a triplet loss to generate the final embedding for retrieval.
    We develop the following variants of this network as: \\
    -- \textbf{Doodle-SingleNet. }We replace the architecture in~\cite{Doodle} with a single branch of VGG-16, which can take data from any domain as input. \\
    -- \textbf{Doodle-SingleNet-w/o Label.} We further modify Doodle-SingleNet and remove the domain-discriminator loss function from the training process, so that it can be applied to any unseen domain data. \\
    -- For fair comparison, we replace the VGG-16 backbone in both these variants with SE-ResNet50 and develop \textbf{Doodle-SE-SingleNet} and \textbf{Doodle-SE-SingleNet-w/o Label} respectively.
    \item \textbf{SAKE~\cite{SAKE}} has a single branch of network, with SE-ResNet50 as backbone. 
    It processes both sketch and image data, augmented with a binary domain-label, and knowledge transfer from a pre-trained \emph{Teacher} network. 
    For comparing with SnMpNet, we develop the following variant of SAKE: \\
    -- \textbf{SAKE-w/o Label. } In this variant, we remove the binary domain-indicator from the training process. \\
    -- As in SAKE~\cite{SAKE}, we perform experiments of this variant with embeddings of different dimensions.
\end{enumerate}
Apart from Doodle-to-search~\cite{Doodle} and SAKE~\cite{SAKE}, we have also compared SnMpNet with \textbf{CVAE}~\cite{CVAE}.
We summarize the comparisons in terms of mAP@200 and Prec@200 in Table~\ref{tab_sketchy}.
We observe the best performance is obtained through the SAKE-model with the domain indicator~(our evaluation)\footnote{SAKE-model trained using the split and evaluation in~\cite{Doodle}}.
Note that this model cannot be used for unseen query domains in UCDR protocol, because of the domain indicator.
We also observe that the performance of both the state-of-the-art approaches, Doodle-to-Search~\cite{Doodle} and SAKE~\cite{SAKE} degrade drastically when either the domain-specific two-branch architecture or the domain-indicator is removed.
SnMpNet outperforms these variants, CVAE and Doodle-to-Search, which justifies its effectiveness.

\subsection{UCDR Evaluation}
We now extend our evaluation for the fully generalized UCDR protocol on DomainNet~\cite{Domainnet}.
Since there is no existing baseline for this, we develop two variants of very closely related works present in literature.
We start with a brief description of these variants. \\
\textbf{Baseline Methods: }
We consider two state-of-the-art approaches developed for related applications, namely
1) EISNet~\cite{eisnet}, which is the SOTA for DG and 
2) CuMix~\cite{Cumix}, the first work to address ZSDG.
Since these have been developed for classification, we make minimal changes in the networks, to address the retrieval task in UCDR.

\textbf{1) EISNet-Retrieval:}
We incorporate a 300-d linear-layer in the classification branch in~\cite{eisnet}, whose output is used as the domain-invariant feature for UCDR.

\textbf{2) CuMix-Retrieval:}
For fair comparison, we use SE-ResNet50 as backbone with a 300-d linear layer on top in~\cite{Cumix} and incorporate the image and feature-level mixing method, as proposed in CuMix. We discuss the details of these modifications in Supplementary. \\
For UCDR, we train with \emph{seen}-class samples from 5-domains, leaving one domain out.
The \emph{unseen} class samples from this \emph{unseen}-domain are used as query for evaluation. 
We evaluate for two configurations of search set, where it contains images from: (a) only \emph{unseen}-classes, and (b) both \emph{seen} and \emph{unseen} classes.
Clearly, (b) is more challenging than (a), because of the scope of greater confusion.
We report the individual results on all 5-domains~(except \emph{Real}) as query, as well as the average retrieval accuracy in Table~\ref{tab_UCDR}.
We observe that for all the approaches, the performance degrades significantly when both seen and unseen classes are present in the search set.
However, SnMpNet outperforms the other baselines by a considerable margin.

\section{Analysis}
\label{sec_analysis}
We analyze the contribution of different components of SnMpNet, and its performance in other retrieval scenarios.

\vspace{1.5mm}

\noindent
{\bf U$^c$CDR-Evaluation on DomainNet: }
\label{subsec_uccdr}
We now present the evaluation of SnMpNet for U$^c$CDR on DomainNet. Here the query domain \emph{Sketch} or \emph{QuickDraw} is \emph{seen}, but the query samples belong to \emph{unseen} classes.
We perform experiments for two configurations of search set as in $UCDR$.
From Table~\ref{tab_UcCDR}, we observe that SnMpNet outperforms the other approaches.

\vspace{1.5mm}

\noindent
{\bf U$^d$CDR-Evaluation on DomainNet:}
Here, for completeness, we evaluate SnMpNet for U$^d$CDR, where the query belongs to a seen class, but from an \emph{unseen} domain.
To construct the query set, we randomly select $25\%$ samples from each of the \emph{seen} classes from \emph{Sketch} domain.
The search set contains images from \emph{seen}-classes.
From Table~\ref{tab_UdCDR}, we observe that SnMpNet significantly outperforms the two strong baselines.
This setup can be considered as a modification of domain generalization problem for the retrieval task.
Here, the knowledge gap between the training and test classes is not present. 
The main challenge for the retrieval model is to address the domain gap because of the unseen query domain.

\begin{table}[ht!]
\footnotesize
\begin{center}
\begin{tabular}{|c|c|c|}
\hline
Method & mAP@200 & Prec@200 \\
\hline
EISNet-retrieval & 0.2210 & 0.1094 \\
CuMix-retrieval & 0.2703 & 0.1224 \\
\textbf{\emph{SnMpNet}} & \textbf{0.3529} & \textbf{0.1657} \\
\hline 
\end{tabular}
\end{center}
\vspace{-4mm}
\caption{U$^d$CDR-evaluation on DomainNet for \emph{unseen} sketch query domain.
The search set contains only seen class real images. 
The models are trained on 5-domains \emph{Real}, \emph{QuickDraw}, \emph{Infograph}, \emph{Painting} and \emph{Clip-art}.}
\label{tab_UdCDR}
\end{table}

\vspace{-2mm}

\noindent
{\bf Ablation Study:}
Now, we analyze the effectiveness of different components of SnMpNet on Sketchy-extended for ZS-SBIR.
We first consider the simplest form of our network (Base N/W), which is SE-ResNet50, trained with the cross-entropy loss evaluated as the cosine-similarity of model output with seen-classes' semantic information~\cite{Glove}~(equation~\eqref{eq_standard_CE}).
The performance of Base N/W, and as each loss component is appended to the base module is summarized in Table~\ref{tab_ablation_sketchy}.
We observe that each component contributes positively to the overall performance.

\begin{table}[ht!]
\footnotesize
\begin{center}
\begin{tabular}{|c|c|c|}
\hline
Proposed Network Variants & mAP@200 & Prec@200 \\
\hline
Base N/W & 0.5218 & 0.4497 \\
Base N/W + $\mathcal{L}_{Sn}~(\kappa=0)$  & 0.5593 & 0.5002\\
Base N/W + $\mathcal{L}_{Sn}~(\kappa=2)$ &  0.5613 & 0.5030 \\
Base N/W + $\mathcal{L}_{CE}^{mix}$ &   0.5252 & 0.4530 \\
Base N/W + $\mathcal{L}_{CE}^{mix}+\mathcal{L}_{Mp}$ &  0.5665 & 0.4989 \\
\textbf{\emph{SnMpNet}} &  \textbf{0.5781} & \textbf{0.5155} \\ 
\hline
\end{tabular}
\end{center}
\vspace{-4mm}
\caption{Ablation study of the proposed SnMpNet framework for ZS-SBIR on Sketchy extended dataset~\cite{CVAE}. }
\label{tab_ablation_sketchy}
\vspace{-4mm}
\end{table}

\begin{figure*}[ht!]
\centering
\includegraphics[width=\textwidth]{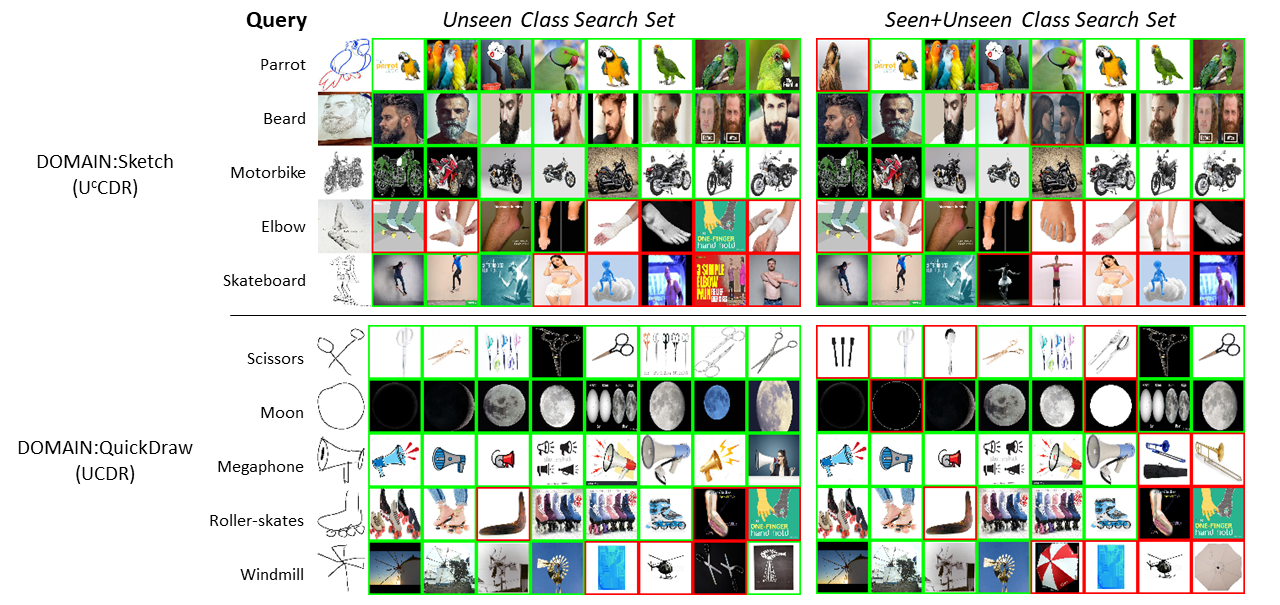}
\caption{Top-8 retrieved Images for UCDR and U$^c$CDR protocols on DomainNet with QuickDraw being the unseen query domain. 
Same query is considered for both the search set configurations. \emph{Green} and \emph{Red} borders indicate correct and incorrect retrievals respectively. (best viewed in color)}
\label{fig:Retrieval_samples_sketch_seen_model}
\vspace{-4mm}
\end{figure*}

\begin{figure}[ht!]
\centering
\includegraphics[width=0.5\textwidth]{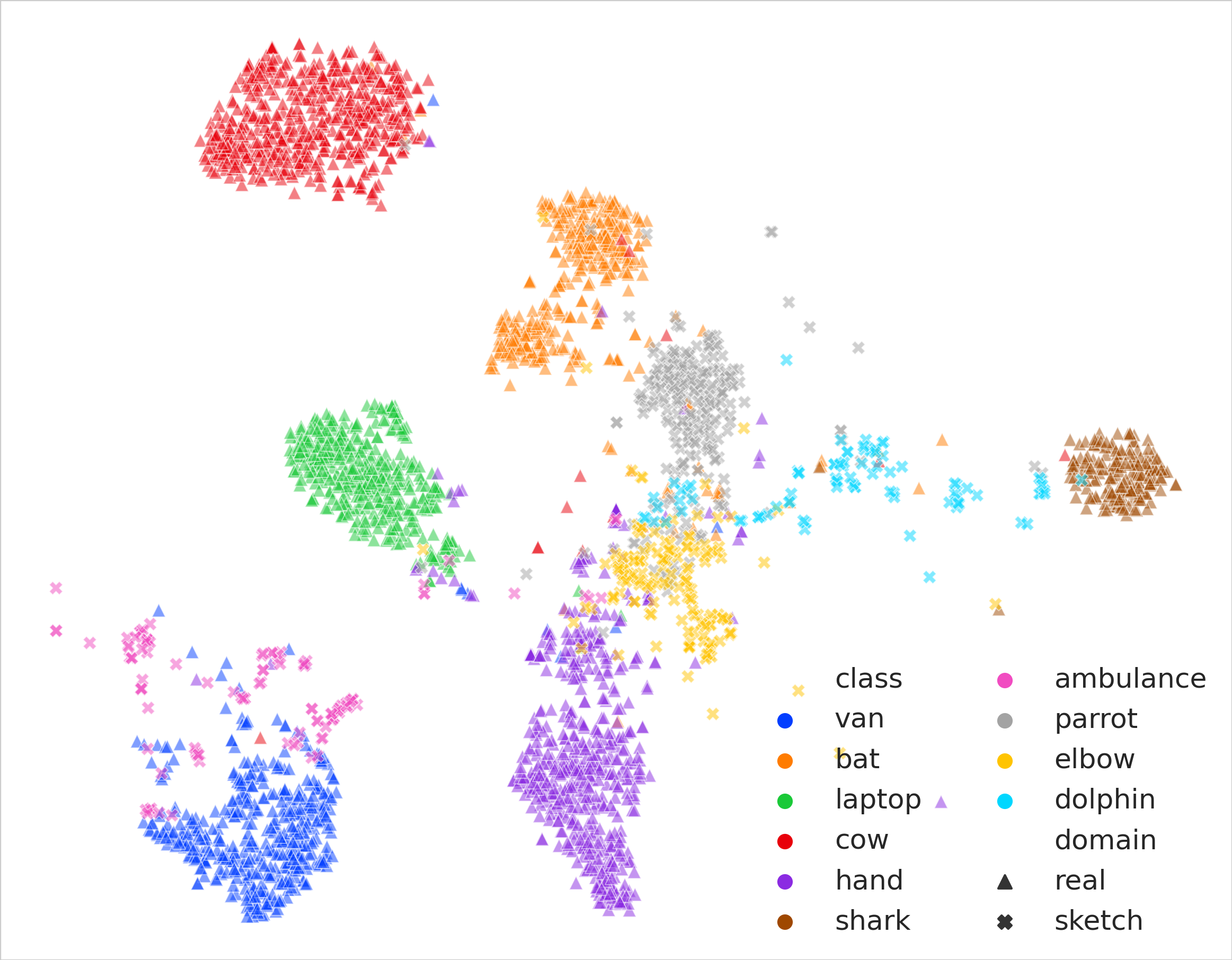}
\caption{t-SNE~\cite{tsne} plot for few randomly selected \emph{seen}~(\emph{van}, \emph{laptop}, \emph{cow}, \emph{hand}, \emph{bat}, \emph{shark}) and \emph{unseen}-classes~(\emph{ambulance}, \emph{elbow}, \emph{parrot}, \emph{dolphin}) in the feature-space using proposed SnMpNet. Here, \emph{Sketch} is \emph{unseen} to the model, while \emph{real} is \emph{seen}. (best viewed in color)}
\label{fig_snmpnet_tsne}
\vspace{-6mm}
\end{figure}

\subsection{Qualitative Results}
\label{sec_analysis_qualititive}
Figure~\ref{fig:Retrieval_samples_sketch_seen_model} shows top-$8$ retrieved images for few queries for UCDR and U$^c$CDR, with QuickDraw as \emph{unseen} domain.
As expected, the results degrade when both \emph{seen} and \emph{unseen} classes are present in the search set.
We also observe that some of the incorrect retrievals are because of shape similarities between classes, like \emph{helicopter} and \emph{windmill}, while some others are due to co-occurrence of different classes in the same image~(\emph{sweater} and \emph{elbow} for \emph{skateboard}).

The t-SNE~\cite{tsne} plot of the feature-space for some randomly selected classes from \emph{seen}~(image) and \emph{unseen}~(sketch) domains is shown in Figure~\ref{fig_snmpnet_tsne}. 
We observe that the \emph{unseen}-classes - namely, \emph{ambulance}, \emph{dolphin}, \emph{parrot}, and \emph{elbow} from the \emph{unseen}-domain sketch are placed in the neighbourhood of the related \emph{seen}-classes - \emph{van}, \emph{shark}, \emph{bat}, and \emph{hand} respectively from \emph{seen}-domain, image, further justifying the effectiveness of the proposed framework. 
\section{Conclusion}
\label{sec_conclusion}
In this work, we proposed a novel framework,  \textbf{SnMpNet} for the Universal Cross-domain Retrieval task.
To the best of our knowledge, this is the first work which can handle query data from unseen classes and unseen domains for retrieval.
In addition to defining the experimental protocol, we also proposed a novel framework, SnMpNet, which introduces two novel losses, {\em Semantic Neighbourhood loss} and {\em Mixture Prediction loss} for the UCDR task.
Extensive experiments and comparisons on two large-scale  datasets, corroborate effectiveness of the proposed SnMpNet. 
\blfootnote{\textbf{Acknowledgements }This work is partly supported through a research grant from SERB, Department of Science and Technology, Govt. of India.}

{\small
\bibliographystyle{ieee_fullname}
\bibliography{egbib}
}

\onecolumn
\appendix

\section{Supplementary materials}
In this supplementary material, we discuss and analyse more experimental evidences which corroborate the effectiveness of the proposed \emph{Semantic Neighbourhood and Mixture Prediction Network}~(SnMpNet).
We start this discussion with the in-depth details of the two baseline methods developed for UCDR. 

\subsection{Implementation Details for UCDR Baseline Methods}
As discussed in the main paper, we modified two recent algorithms from the field of Domain Generalization and Zero-shot Domain Generalization so that they can serve as the baseline retrieval models for UCDR. Both these baselines are subjected to the same train-time augmentations as SnMpNet.\\ \\
\textbf{EISNet-Retrieval. }
We replace the backbone network of EISNet~\cite{eisnet} (originally ResNet50) with SE-ResNet50~(same as our SnMpNet).
On top of the last layer of this backbone, we insert a 300-d linear layer, which is connected to the classification branch~(the branch with the CE-loss in~\cite{eisnet}) of the network.
We further modify this CE-loss in~\cite{eisnet}, and train the model with the semantic-embedding similarity based CE-loss, described in equation~\eqref{eq_standard_CE}.
This modification in CE-loss accounts for the semantic information used in proposed SnMpNet~(which was not present in~\cite{eisnet}'s original architecture) and thus it is fair to compare the performance of this modified EISNet-retrieval model with proposed SnMpNet. 
We leave the other branches of EISNet-architecture~\cite{eisnet} and the corresponding losses unchanged. \\ \\
\textbf{CuMix-Retrieval. }
Similar to the previous case, here also we use SE-ResNet50 as the backbone network.
In addition, we leverage the image-level and feature-level cross-domain and cross-category mixing of samples, as introduced in~\cite{Cumix}; and thus we refer to this model as CuMix-Retrieval.
To account for category-discrimination of these mixed-up samples in the learned feature-space, we first apply the mix-up classification loss~$\mathcal{L}_{CE}^{mix}$~(equation~\eqref{eq_mix_CE} in main paper) on samples mixed-up at the image-level.
Towards that goal, we employ a 300-d linear layer on top of the last layer of the SE-ResNet50 backbone, so that the output of this 300-d layer can be used to compute the semantic similarities with class-embeddings, which are required for $\mathcal{L}_{CE}^{mix}$.
To make use of the feature-level mixing, we generate $\mathbf{g^\ast} = \alpha \mathbf{g}_i^{c,d} + (1-\alpha)[\beta \mathbf{g}_j^{p,d} + (1-\beta)\mathbf{g}_k^{r,n}]$, at the output of the backbone network; and consequently pass it through the above-mentioned 300-d layer to obtain the final feature $\mathbf{f}^*$.
We now re-compute $\mathcal{L}_{CE}^{mix}$ as $\mathcal{L}_{CE}^{mix-feat}$ with respect to these $\mathbf{f}^*$-features.
Thus, the final loss becomes, $\mathcal{L}_{CuMix-Retrieval} = \mathcal{L}_{CE} + \omega_1\mathcal{L}_{CE}^{mix} + \omega_2\mathcal{L}_{CE}^{mix-feat}$, where $\mathcal{L}_{CE}$ is the standard cross-entropy loss computed for original~(not generated through mix-up) samples, and $\omega_1$ and $\omega_2$ are two experimental hyper-parameters.
Thus, this CuMix-Retrieval model now has access to the additional knowledge introduced by the class-name embeddings and the same backbone network as SnMpNet; and hence it is fair to compare its performance with SnMpNet for UCDR.

It can be observed that the final features of test samples~(both query and search set data) obtained from these models are 300-dimensional. Thus, in absence of any existing methods in literature, these baseline models serve as the logical and fair competitors of proposed SnMpNet. \\

\noindent Now, we move on to the additional experimental analysis presented in the following section.

\subsection{Experiments and Analysis contd.}
In this section, we explore the robustness of the proposed model by analyzing it for different variations of protocols. \\ \\
\textbf{1) U$^{c}$CDR Evaluation on additional datasets. }
We discussed the performance of proposed SnMpNet for U$^c$CDR-protocol in Table~\ref{tab_UcCDR}, and compared it with existing ZS-SBIR methods in Table~\ref{tab_sketchy}, ZS-SBIR being a special form of the same protocol, where sketch and real images are used as query and search-set domains, respectively.
We extend this analysis further on two other challenging datasets: 1) Sketchy-extended~\cite{Sketchy}, with split proposed in~\cite{ZSIH} - randomly chosen $25$-classes are used as \emph{unseen} test classes and out of the rest $100$-classes, $90$ are used for training and $10$ used for validation; and 2) TU-Berlin\footnote{M. Eitz, J. Hayes and M. Alexa, How do humans sketch objects?, \emph{SIGGRAPH}, 2012}, with split followed in~\cite{ZSIH}\cite{SEM_PCYC} etc. - out of total $250$-classes, randomly chosen $30$-classes~(with at least $400$-images per category) are kept for testing, and $220$-classes~($200$ for training and $20$ for validation) are used as \emph{seen}-classes.

\begin{figure}[htbp]
\centering
\includegraphics[scale=0.6]{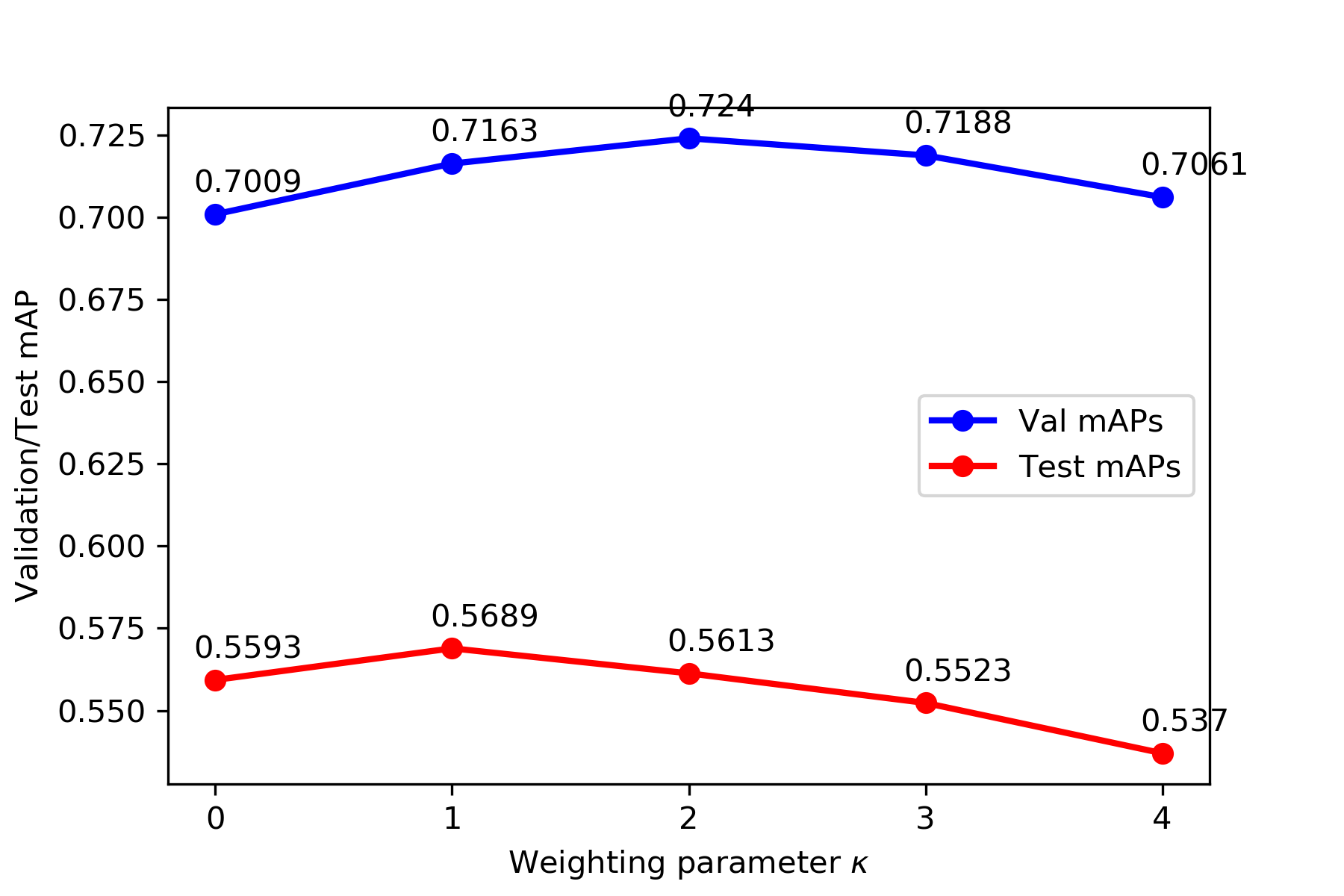}
\caption{Effect of $\kappa$ on validation and test mAP@200 for Sketchy extended split~\cite{CVAE}.}
\label{fig_kappa_effect}
\end{figure}

The results are summarized in Table~\ref{tab_sketchy_tuberlin_random_splits}. 
\begin{table}[htbp]
\small
\begin{center}
\begin{tabular}{|c|c|c|c|c|c|c|c|}
\hline
& {\multirow{2}{*}{Method}} & \multirow{2}{*}{Backbone Network} & \multirow{2}{*}{output dim.} & \multicolumn{2}{c|}{TU-Berlin extended} & \multicolumn{2}{c|}{Sketchy extended} \\
\cline{5-8}
& & & & mAP@all & Prec@100 & mAP@all & Prec@100 \\
\hline
\multirow{4}{*}{Existing SOTA} & SEM-PCYC~\cite{SEM_PCYC}~(CVPR'19) & VGG-16 & 64 & 0.297 & 0.426 & 0.349 & 0.463 \\
& Style-guide~\cite{StyleGuide}~(TMM'20) & VGG-16 & 200 & 0.2543 & 0.3551 & 0.3756 & 0.4842 \\
& SAKE-512~\cite{SAKE}~(ICCV'19) & SE-ResNet50 & 512 & 0.475 & 0.599 & 0.547 & 0.692 \\
& SAKE-512 (our evaluation) & SE-ResNet50 & 512 & 0.3468 & 0.5225 & 0.4690 & 0.6665 \\
\hline
\multirow{2}{*}{SAKE-Variants} & SAKE-512-w/o Label & SE-ResNet50 & 512 & 0.3314 & 0.5052 & 0.3599 & 0.5216\\
& SAKE-300-w/o Label & SE-ResNet50 & 300 & 0.3233 & 0.4959 & 0.3312 & 0.4841 \\
\hline
\multicolumn{2}{|c|}{\textbf{\emph{SnMpNet}}} & SE-ResNet50 & 300 & \textbf{0.3568} & \textbf{0.5226} & \textbf{0.4412} & \textbf{0.5887}\\
\hline
\end{tabular}
\end{center}
\caption{Peformance comparison for ZS-SBIR on Sketchy extended and TU-Berlin.}
\label{tab_sketchy_tuberlin_random_splits}
\end{table}
We use mAP@all and Prec@100 as the evaluation metrics.
For SEM-PCYC~\cite{SEM_PCYC} and StyleGuide~\cite{StyleGuide}, we compare with the reported results in the respective papers.
On the other hand, we use the implementation provided by the authors of~\cite{SAKE}, but unfortunately were unable to reproduce their reported numbers.
Thus, we present the retrieval accuracies obtained by us as SAKE~(our evaluation) in Table~\ref{tab_sketchy_tuberlin_random_splits}, as well as the originally reported mAP-values for reference.
Out of the SOTA-methods listed in the table, owing to the nature of the algorithms, it is feasible to create domain-independent variants only for SAKE~\cite{SAKE}. 
Thus, similar to Table~\ref{tab_sketchy} in the main paper, here also we evaluate the two SAKE-variants for comparison. 
Following the same pattern as before, we observe that SAKE's performance deteriorates for both of these selected datasets when the domain-indicator is removed.
Our SnMpNet outperforms both of these variants, as well as SEM-PCYC~\cite{SEM_PCYC} and Style-guide~\cite{StyleGuide} for both the datasets. 
Additionally, it has a superior performance even over the original SAKE-model with the domain indicator, on TU-Berlin, further validating its suitability for the UCDR protocol. \\ \\
\textbf{2) U$^d$CDR-Evaluation for QuickDraw. }
Previously in Table~\ref{tab_UdCDR} in the main paper, we have evaluated and compared SnMpNet with the two retrieval baselines, EISNet-retrieval and CuMix-retrieval, on the $U^{d}CDR$ protocol using \emph{Sketch} as the unseen domain. 
For a more exhaustive analysis and completeness,  we now repeat the same experiment with Quickdraw as the unseen domain.
We construct the query set here with 10\% of available samples, selected randomly, from each of the seen classes in QuickDraw. The search set again contains the \emph{seen}-class RGB images as before.
\begin{table}[htbp]
\small
\begin{center}
\begin{tabular}{|c|c|c|}
\hline
Method & mAP@200 & Prec@200 \\
\hline
EISNet-retrieval & 0.0637 & 0.0309 \\
CuMix-retrieval & 0.0648 & 0.0298 \\
\textbf{\emph{SnMpNet}} & \textbf{0.1077} & \textbf{0.0509} \\
\hline 
\end{tabular}
\end{center}
\caption{U$^d$CDR-evaluation on DomainNet for \emph{unseen} QuickDraw query domain. The search set contains only seen class real images. The models are trained on 5 domains - \emph{Real}, \emph{Sketch}, \emph{Infograph}, \emph{Painting} and \emph{Clip-art}. }
\label{tab_UdCDR_quickdraw}
\end{table}
We summarize the retrieval results in Table~\ref{tab_UdCDR_quickdraw} and observe that SnMpNet outperforms the two baselines here as well.  \\ \\
\textbf{3) Effect of Multi-domain Training data. }
Here we explore the effect of using training data from more than 2 domains to address the cross-domain retrieval.
Towards this end, we train SnMpNet using only 2-domains~(M=2)~(\emph{Sketch}/\emph{QuickDraw} and \emph{Real}) and observe the retrieval performance for UCDR and U$^{c}$CDR protocols.
The results are summarized in Table~\ref{tab_multi_domain} for two configurations of the search set - a) when only \emph{unseen} classes are present in the search set and b) the search set contains samples from both \emph{seen} and \emph{unseen} classes.
We also mention SnMpNet's performance using all 5-training domains (a common practice in DG) in Table~\ref{tab_multi_domain} for ease of comparison.
Importantly, we observe that the performance boost on using the three auxiliary domains is more pronounced for UCDR than U$^{c}$CDR.
\begin{table}[htbp]
\begin{center}
\begin{tabular}{|c|c|c|c|c|c|c|}
\hline
Protocol & Query Domain & Training Domains  & \multicolumn{2}{c|}{\emph{Unseen}-class Search Set} & \multicolumn{2}{c|}{\emph{Seen}+\emph{Unseen}-class Search Set} \\
\cline{4-7}
& & & mAP@200 & Prec@200 & mAP@200 & Prec@200 \\
\hline

\multirow{8}{*}{$UCDR$} & \multirow{4}{*}{\emph{QuickDraw}} & \emph{Sketch}, \emph{Real} & 0.1540 & 0.1138 & 0.1332 & 0.0972 \\
\cline{3-7}
& & \emph{Sketch}, \emph{Real},  & \multirow{3}{*}{0.1736} & \multirow{3}{*}{0.1284} & \multirow{3}{*}{0.1512} & \multirow{3}{*}{0.1111}  \\ 
& & \emph{Infograph}, \emph{Painting},  & & & & \\
& &  \emph{Clip-art}  & & & & \\

\cline{2-7}

& \multirow{4}{*}{\emph{Sketch}} & \emph{QuickDraw}, \emph{Real} & 0.2490 & 0.1953 & 0.2188 & 0.1741 \\
\cline{3-7}
& & \emph{QuickDraw}, \emph{Real},  & \multirow{3}{*}{0.3007} & \multirow{3}{*}{0.2432} & \multirow{3}{*}{0.2624} & \multirow{3}{*}{0.2134}  \\ 
& & \emph{Infograph}, \emph{Painting},  & & & & \\
& &  \emph{Clip-art}  & & & & \\

\hline

\multirow{8}{*}{$U^cCDR$} & \multirow{4}{*}{\emph{Sketch}} & \emph{Sketch}, \emph{Real} & 0.4163 & 0.3455 & 0.3696 & 0.3066 \\
\cline{3-7}
& & \emph{Sketch}, \emph{Real},  & \multirow{3}{*}{0.4221} & \multirow{3}{*}{0.3496} & \multirow{3}{*}{0.3767} & \multirow{3}{*}{0.3109}  \\ 
& & \emph{Infograph}, \emph{Painting},  & & & & \\
& &  \emph{Clip-art}  & & & & \\

\cline{2-7}

& \multirow{4}{*}{\emph{QuickDraw}} & \emph{QuickDraw}, \emph{Real} & 0.2763 & 0.2181 & 0.2215 & 0.1832 \\
\cline{3-7}
& & \emph{QuickDraw}, \emph{Real},  & \multirow{3}{*}{0.2888} & \multirow{3}{*}{0.2314} & \multirow{3}{*}{0.2366} & \multirow{3}{*}{0.1918}  \\ 
& & \emph{Infograph}, \emph{Painting},  & & & & \\
& &  \emph{Clip-art}  & & & & \\

\hline

\end{tabular}
\end{center}
\caption{Effect of training data from multiple-domains on Retrieval Performance of \emph{SnMpNet}.}
\label{tab_multi_domain}
\end{table}
Thus, we can infer that the information from auxiliary training domains enhances the model's generalization abilities, especially for a new \emph{unseen} domain. \\ \\
\textbf{4) Effect of weighting parameter $\kappa$. }
In the main paper, we propose a novel semantic neighbourhood loss $\mathcal{L}_{Sn}$, which implements a strict-to-relaxed weighting scheme based on how closely another class is related to the class of the current sample. 
It can be observed from equation~\eqref{eq_wtvec_defn}, that the most important hyper-parameter we use in the training process of SnMpNet is $\kappa$, which effectively controls this above-said weighting.
In Figure~\ref{fig_kappa_effect}, we perform a simple experiment to observe the effect of this hyper-parameter $\kappa$ in proposed $\mathcal{L}_{Sn}$.
To this end, we vary $\kappa$ over a range $0 \leq \kappa \leq 4$ and observe the retrieval performance~(mAP@200) on the validation set data, of a primitive variant of \emph{SnMpNet} - Base N/W + $\mathcal{L}_{Sn}$ (Table~\ref{tab_ablation_sketchy} in the main paper). 
We again perform this experiment for ZS-SBIR protocol on the Sketchy-extended dataset~\cite{CVAE}.

From equation~\eqref{eq_wtvec_defn} in the main paper, it follows that $\kappa=0$ corresponds to having equal weights for all classes, i.e. on all elements of the difference, $||\mathbb{D}(\mathbf{f}_i^{c,d}) - \mathbb{D}_{gt}(\mathbf{f}_i^{c,d})||^2$.
For $\kappa\neq 0$, $\mathbf{w}(c)_c = 1$ and $\mathbf{w}(c)_k = e^{-\kappa}$, where $k$ is the index of the most dissimilar class of $c$, in terms of their semantic distance $D(\mathbf{a}^c, \mathbf{a}^k)$.
For any other class pair $(c, j)$ ($j \neq k$), $e^{-\kappa} < \mathbf{w}(c)_j < 1$, thereby enforcing the strict-to-relaxed criterion for preserving relative distances between classes.

As shown before in Table~\ref{tab_ablation_sketchy}, we select the model corresponding to $\kappa=2$ for testing based on its highest retrieval accuracy on the validation set.
For completeness, the variation of test set retrieval performance with $\kappa$ is also plotted in the same figure. 
\subsection{Qualitative Analysis contd.}
\begin{figure}[htbp]
\centering
    \begin{subfigure}[h!]{\textwidth}
        \centering
        \includegraphics[width=\textwidth]{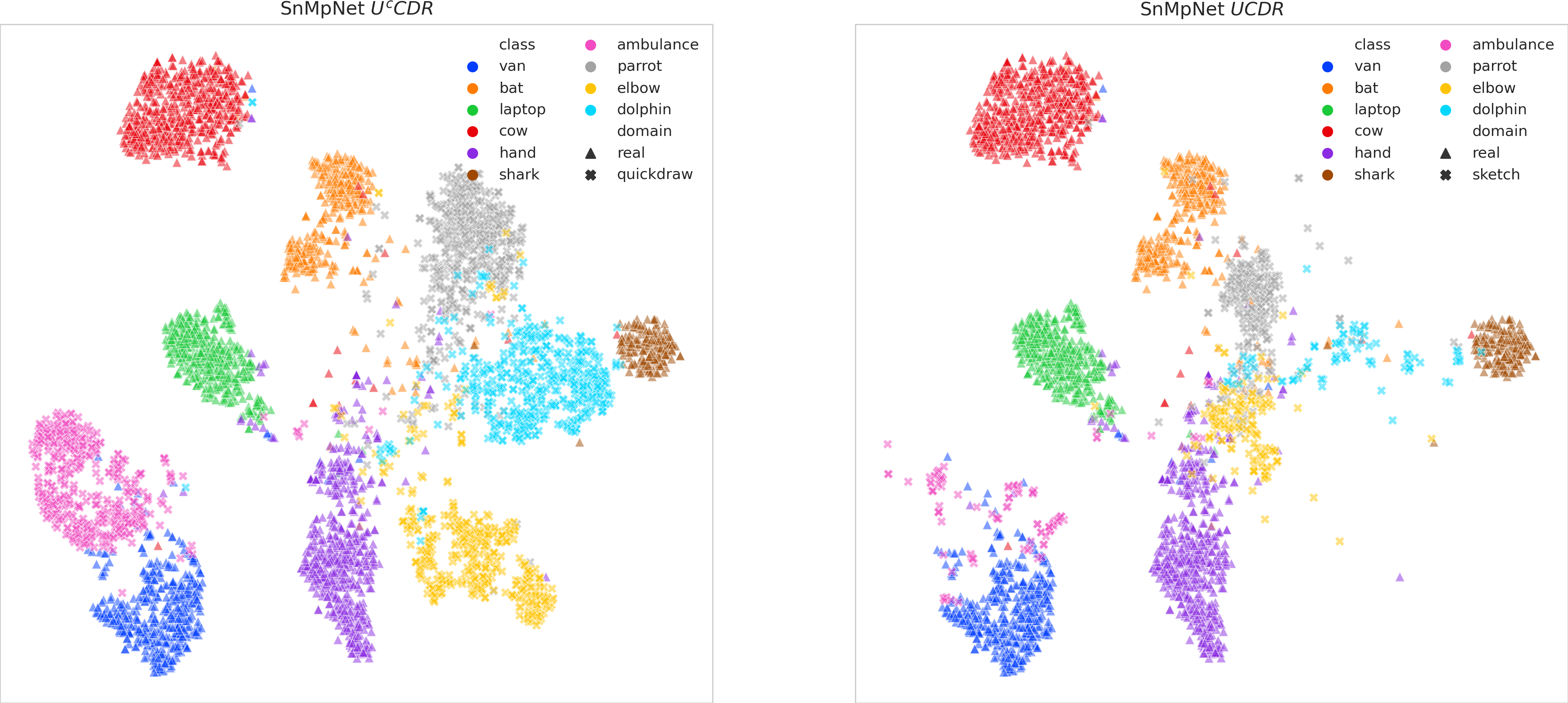}
    \end{subfigure}
    \begin{subfigure}[h!]{\textwidth}
        \includegraphics[width=\textwidth]{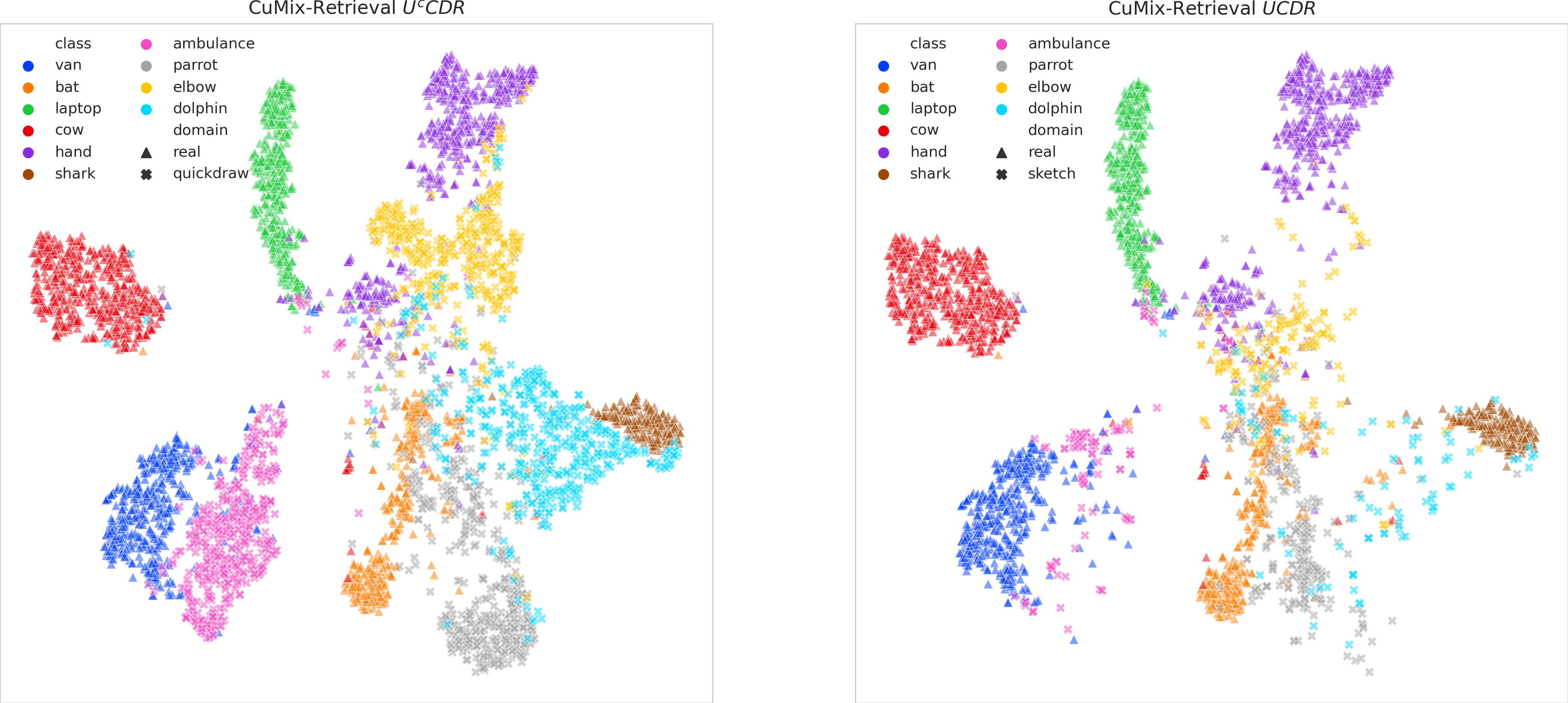}
    \end{subfigure}
\caption{\color{black}t-SNE plots for UCDR and U$^c$CDR protocols with CuMix-Retrieval and proposed SnMpNet. Here \emph{Sketch} is unseen to the models, while \emph{QuickDraw} and \emph{Real} are seen. (best viewed in color)}
\label{fig:tsne_cumix_snmpnet}
\end{figure}

\begin{figure}[ht!]
\centering
    \begin{subfigure}[h!]{\textwidth}
        \centering
        \includegraphics[width=\textwidth]{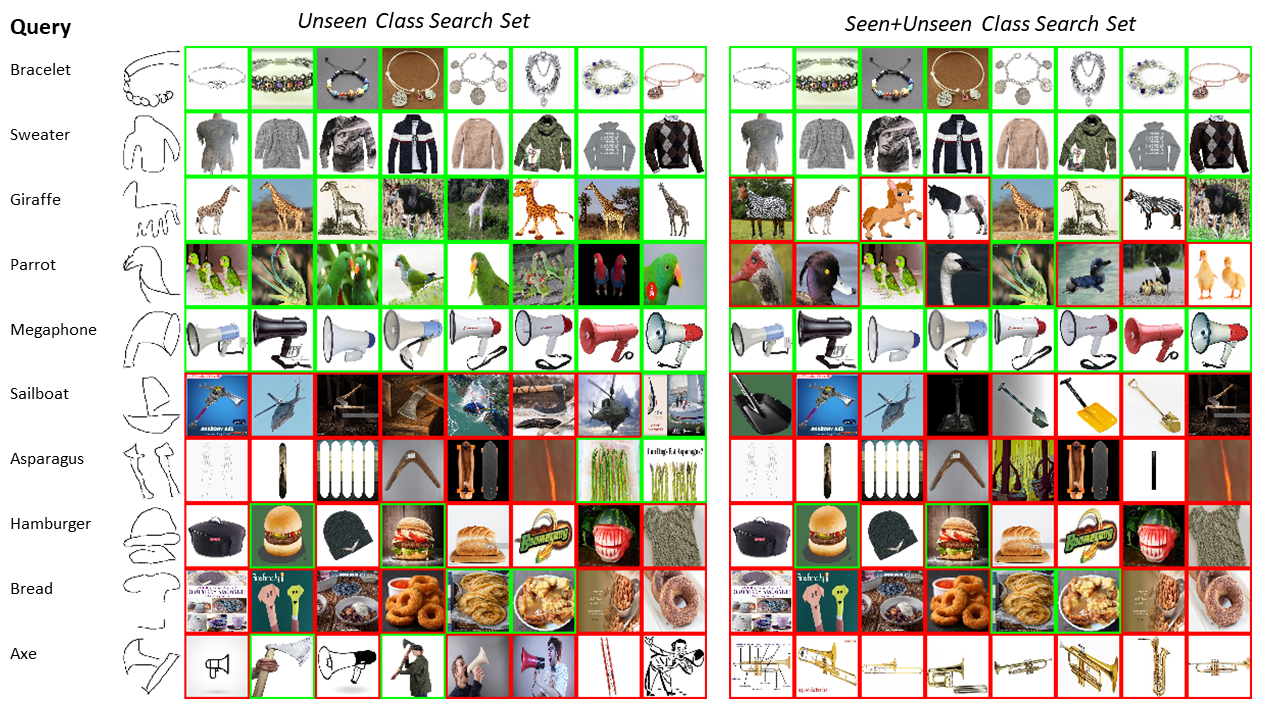}
        \caption{\color{black}$U^{c}CDR$ for QuickDraw}
    \end{subfigure}
    \begin{subfigure}[h!]{\textwidth}
        \includegraphics[width=\textwidth]{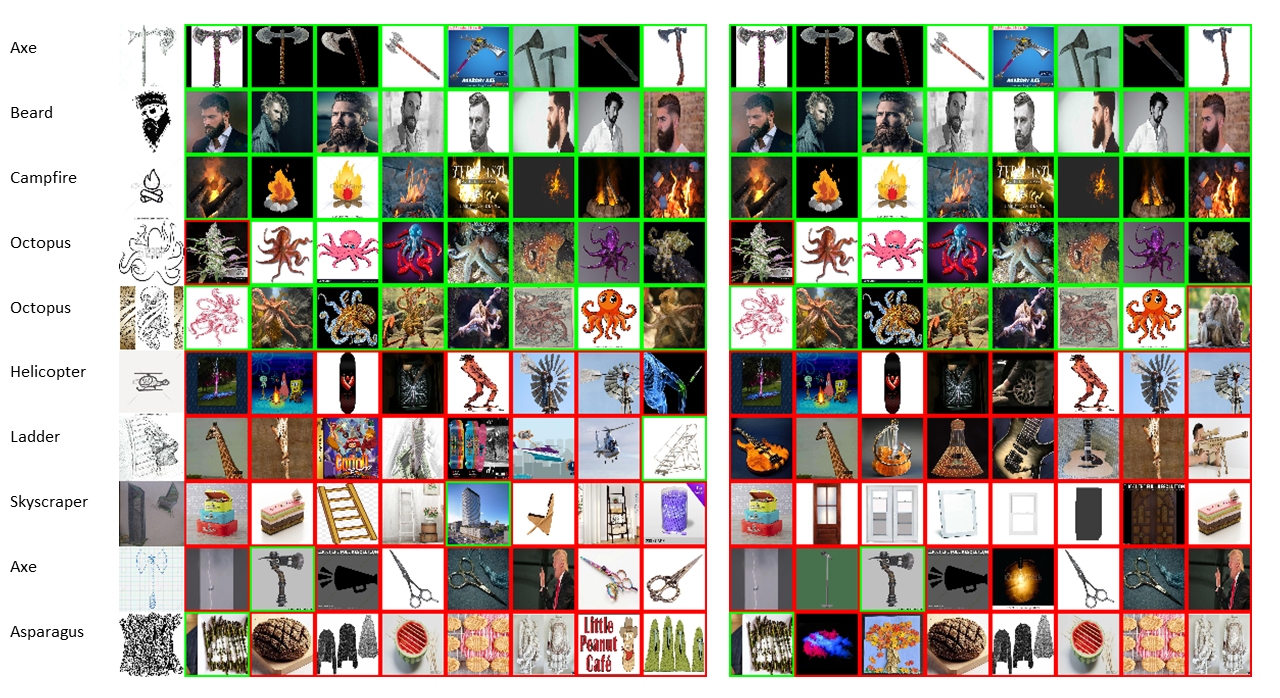}
        \caption{$UCDR$ for Sketch}
    \end{subfigure}
\caption{Top-8 Retrieved Images for UCDR and U$^c$CDR protocols on DomainNet with Sketch being the unseen query domain. Same query is considered for both the search set configurations. \emph{Green} and \emph{Red} borders indicate correct and incorrect retrievals respectively. (best viewed in color)}
\label{fig:Retrieval_samples_quickdraw_seen_model}
\end{figure}

Here we further analyse SnMpNet to observe the feature-space and retrieved samples in details. \\ \\
\textbf{1) Visualization of the feature-space. }
We further visualize the learned feature-space through SnMpNet using a t-SNE~\cite{tsne} plot.
Towards that goal, we train SnMpNet with samples from seen classes, which belong to the domains - Real, Quickdraw, Clip-art, Painting and Infograph.
For visualization, we pick the same $10$-categories as in Figure~\ref{fig_snmpnet_tsne}.
We project features from these selected seen and unseen classes onto the feature-space demonstrated in Figure~\ref{fig:tsne_cumix_snmpnet}.
Moreover, we project features from both Quickdraw (seen domain, protocol U$^c$CDR in figure) and Sketch (unseen domain, protocol UCDR in figure).
For comparative analysis, we also simulate the feature-space using the CuMix-retrieval algorithm and present the visualizations in the same figure.
We can clearly observe better categorical distinction using SnMpNet over its close-competitor baseline CuMix-Retrieval.
Although CuMix-Retrieval is able to place the 4 \emph{unseen}-classes in the neighborhood of the 4 related \emph{seen}-classes, it is not able to separate the object classes into clear and well-separated clusters for 5 out of the 10 classes. 
Only \emph{van}, \emph{ambulance}, \emph{cow}, \emph{laptop}, and \emph{shark} have easily distinguishable cluster maps. 
SnMpNet, on the other hand, succeeds at both the above objectives and creates near-distinct cluster boundaries for all 10 object classes. \\ \\
\textbf{2) Sample Retrieval Results. }
In continuation to Figure~\ref{fig:Retrieval_samples_sketch_seen_model} in the main paper, we provide some additional retrieval results in Figure~\ref{fig:Retrieval_samples_quickdraw_seen_model}.
In contrast to Figure~\ref{fig:Retrieval_samples_sketch_seen_model}, here we use \emph{Sketch} as the unseen domain for UCDR, and report the top-8 retrieved images against each query for two configurations of the search set, as before.
We have also presented sample retrieved images for a few randomly selected sample queries from a seen domain, Quickdraw, for complete analysis of this instance of trained SnMpNet.
Thus these results correspond to the protocol U$^c$CDR.
We observe the similar \emph{mistake}-pattern as in Figure~\ref{fig:Retrieval_samples_sketch_seen_model}, i.e. the model struggles when the source \emph{queries} (\emph{sketch} or \emph{quickdraw}) are ambiguous, poorly drawn or lack sufficient detail pertaining to the category. 
This results in retrieval of images from categories, which are either semantically related, or unrelated, but have high shape resemblance with the query.
This is observed in several examples: \emph{axe} being retrieved for \emph{sailboat}; \emph{finger}, \emph{skateboard}, \emph{tornado} for \emph{asparagus}; \emph{lightning}, \emph{megaphone} for \emph{axe}; \emph{windmill}, \emph{campfire} for \emph{helicopter}; \emph{cake}, \emph{suitcase}, \emph{ladder} for \emph{skyscraper} etc. 
In the more challenging search set configuration containing both \emph{seen} and \emph{unseen} classes, incorrect retrievals are from more visually similar or closely related categories. 
For example, \emph{bird}, \emph{duck}, \emph{swan}, \emph{penguin} are retrieved for \emph{parrot}; \emph{zebra}, \emph{horse} for \emph{giraffe}; \emph{monkey} for \emph{octopus}; \emph{door} for \emph{skyscraper} etc. \\

Thus, we have analyzed the proposed model SnMpNet for a wide variation of experimental protocols.
We have also explained the hyper-parameters associated with our model for better understanding.
Now, we finally conclude with a brief discussion on how the proposed $UCDR$ protocol  is different from other related works in literature.

\subsection{Difference between proposed UCDR and other related works}
Here, we discuss the differences between SnMpNet and \cite{OCDVS}.
A open cross-domain visual search protocol has been proposed in~\cite{OCDVS}, which is significantly different from the traditional cross-domain data retrieval, which addresses the problem of retrieving data from one fixed target domain, and relevant to the query from another fixed source domain.
This newly proposed protocol is again significantly different from our UCDR.
We summarize the differences here.
\begin{enumerate}
\item Open cross-domain visual search~\cite{OCDVS} proposes the cross-domain search among any two domains, provided they have been used during training.
In contrast, proposed UCDR focuses on cross-domain retrieval scenario, when the query domain is not \emph{seen} during training.
\item Proposed SnMpNet uses multiple domains of data~(more than two) for training, as in~\cite{OCDVS}; however SnMpNet processes all domains through one single network~(feature-extractor + classifier), instead of the separate domain-specific prototypical networks that learn a common semantic space in~\cite{OCDVS}.
This results in significant decrease in the number of trainable parameters and model complexity.
\item \cite{OCDVS} requires separate learning of a new semantic mapping function, whenever a new source / target domain emerges.
Since the proposed model in~\cite{OCDVS} requires a-priori knowledge about the query and target domains, it cannot be used for UCDR protocol, without additional training.
In contrast, SnMpNet can be seamlessly extended to the proposed multi-domain query / target conditions, proposed in~\cite{OCDVS}.
\end{enumerate}
It can be noticed that the focus of our work is more towards the generalization ability of the network for \emph{unseen} classes and \emph{unseen} domains, whereas~\cite{OCDVS} works towards generalizing retrieval in case of any query-search set pairs from \emph{seen}-domains.
Thus our work is significantly different from~\cite{OCDVS}.

\end{document}